\pdfoutput=1

\documentclass[11pt]{article}

\usepackage[final]{acl}

\usepackage{times}
\usepackage{latexsym}

\usepackage[T1]{fontenc}

\usepackage[utf8]{inputenc}

\usepackage{microtype}

\usepackage{inconsolata}
\usepackage{algorithm}

\usepackage{times}
\usepackage{tikz,pgfplots}
\usepackage{latexsym}
\usepackage{rotating,tabularx,siunitx}
\usepackage{stmaryrd}
\usepackage{footnote}
\usepackage{comment}
\usepackage{graphicx}
\usepackage{todonotes}
\usepackage{paralist}
\usepackage{amsthm}

\theoremstyle{plain}

\newtheorem{definition}{Definition}

\usepackage{url}            
\usepackage{booktabs}       
\usepackage{amsfonts}       
\usepackage{nicefrac}       
\usepackage{microtype}      
\usepackage{xcolor}         
\usepackage{graphicx}
\usepackage{amsmath}
\usepackage{amssymb}
\usepackage{multirow}
\usepackage{color}
\usepackage{algorithm} 
\usetikzlibrary{plotmarks}
\usepackage{algpseudocode} 
\usepackage[normalem]{ulem}
\useunder{\uline}{\ul}{}
\title{Enhancing Transformers for Generalizable First-Order Logical Entailment}
\author{Tianshi Zheng\thanks{~~Equal Contribution}$^{\textbf{1}}$, Jiazheng Wang\footnotemark[1]$^{\textbf{2}}$, Zihao Wang$^{\textbf{1}}$, Jiaxin Bai$^{\textbf{1}}$, Hang Yin$^{\textbf{3}}$,  Zheye Deng$^{\textbf{1}}$, \\ \bf Yangqiu Song$^{\textbf{1}}$, Jianxin Li$^{\textbf{2}}$ \\
\\
  $^{\textbf{1}}$Department of Computer Science and Engineering, HKUST, Hong Kong SAR, China\\
  $^{\textbf{2}}$Department of Computer Science and Engineering, Beihang University, Beijing, China\\
    $^{\textbf{3}}$Department of Mathematical Sciences, Tsinghua University, Beijing, China\\
  \texttt{tzhengad@connect.ust.hk}\\
}

\begin{document}

\maketitle

\begin{abstract}

Transformers, as the fundamental deep learning architecture, have demonstrated great capability in reasoning. This paper studies the generalizable first-order logical reasoning ability of transformers with their \textit{parameterized} knowledge and how to improve it.
Transformers' capability of first-order reasoning is further captured by whether they can conduct first-order logical entailment, which is quantitatively measured by their performance in answering knowledge graph queries.
We establish the connections between (1) two types of distribution shifts studied in out-of-distribution generalization and (2) unseen knowledge and query settings discussed in the task of knowledge graph query answering, which makes it possible to characterize the fine-grained generalizability. Results on our comprehensive dataset showed that transformers \textit{outperform} previous methods designed particularly for this task and provided detailed empirical evidence about the impact of the input query syntax, token embedding, and transformer architectures on their reasoning capability. Interestingly, our results revealed the mismatch of positional encoding and other design choices of transformer architectures in previous practices. Motivated by this, we propose \textbf{TEGA}, a logic-aware architecture that significantly improves the performance in generalizable first-order logical entailment.
\end{abstract}

%

\section{Introduction}


As a fundamental architecture in deep learning, transformers possess strong reasoning capabilities on various tasks, including arithmetic reasoning~\cite{saxton2019analysing,hendrycks2021measuring}, symbolic reasoning for first-order logic rules~\cite{dehghani2019universal,lample2019deep}, set-theoretic operations \cite{barrett2018measuring}, and theorem proving \cite{polu2020generative}. Besides, transformers have also demonstrated proficiency in logical reasoning over natural language~\cite{han2022folio,tian-etal-2021-diagnosing}. 
To distinguish whether transformers conduct the reasoning rather than fitting the data distribution, recent studies in natural language reasoning further measure the capabilities of transformers for out-of-demonstration samples~\citep{saparov2023testing}. However, their discussions are limited in two aspects: (1) they only concern the reasoning ability of transformers with \textit{in-context} knowledge, and (2) they fail to elicit the connection of out-of-demonstration samples with two distribution shifts~\citep{moreno2012unifying} for the study of out-of-distribution generalization.

In this paper, we take a step further to understand the generalizable reasoning capability of transformers. What sets us apart from~\citep{saparov2023testing} is that (1) our investigation covers the transformer reasoning with \textit{parameterized knowledge}, which suits many scenarios when the related knowledge is not explicit for users and only questions are given, and (2) we further realize two distribution shifts~\citep{moreno2012unifying} in our first order reasoning tasks, which boils down to the process of first-order logical entailment, where we verify whether a first-order sentence (conclusion) is the logical consequence of a set of known first-order sentences (premises)  or not~\citep{marker2002model}.



Specifically, we study the first-order logical entailment with Knowledge Graphs (KG), leading to the widely discussed task of knowledge graph query answering. In our setting, knowledge in KGs is parameterized in models and forms implicit premises. (\S\ref{sec:preliminaries}) The process of logical entailment occurs when identifying the answer to a logical query. Therefore, the two distribution shifts -- concept shift and covariant shift -- are naturally credited to the unobserved knowledge and the unseen query types in our experimental settings. (\S\ref{sec:shifts}) For better evaluation, we build our own benchmark with fifty-five types of logical queries over three knowledge graphs (FB15k~\cite{NIPS2013_1cecc7a7}, FB15k-237~\cite{toutanova-chen-2015-observed}, and NELL995~\cite{10.5555/2898607.2898816}), including all existing features discussed in recent literature~\citep{yin2023rethinking}. Our benchmark results suggest that transformers can handle first-order logical entailment even compared to the methods particularly designed for this task. (\S\ref{sec:benchmark})

We conducted extensive experiments to characterize the impact of three main aspects of solving logical entailment on reasoning capability: the query syntax, the learning of token embedding, and the transformer architecture. Our results provide strong empirical clues on improving transformers' reasoning capability by choosing the proper formal language syntax, positional encodings, semantics in pre-trained KG embeddings, and inductive biases in transformer architectures. (\S\ref{sec:experiments_and_analyses}) 

Interestingly, our results demonstrated the superior performance and generalizability of relative positional encoding (RPE) over traditional absolute positional encoding (APE). However, previous studies have only proposed inductive biases within the APE setting, which we prove ineffective in the RPE setting. To fill this gap, we propose \textbf{TEGA} (\textbf{T}ransformer \textbf{E}ncoder with \textbf{G}uided \textbf{A}ttention), a novel modeling methodology that facilitates effective reasoning with logic-aware guidance in self-attention. Our study shows that TEGA substantially improves the performance and generalizability upon transformers under the RPE setting.\,(\S\ref{sec:tega}) 

Overall, this work makes three key technical contributions:

\begin{enumerate}
    \item \textbf{Benchmark.} We conducted \textbf{comprehensive benchmarking} of transformer architectures against baseline models for knowledge graph query answering, using a new dataset that incorporates \textit{two distribution shifts}.
    
    \item \textbf{Exploration.} We systematically investigated \textbf{critical design choices} throughout the modeling process, including \textit{query syntax and permutation}, \textit{token embedding}, and \textit{transformer architecture designs}.
    
    \item \textbf{Method.} We introduced \textbf{TEGA}, a novel architecture that improves logical entailment performance and generalizability through two inductive biases. We validated its effectiveness through extensive experiments and ablations.
\end{enumerate}

Our code and data can be found at \hyperref[https://github.com/HKUST-KnowComp/TEGA]{\texttt{https://github.com/HKUST-KnowComp/TEGA}}.

\section{Preliminaries}
\label{sec:preliminaries}
This section briefly introduces First-Order (FO) logical entailment and its restricted application in knowledge graph query answering.
By revealing the connection, we show that the key ability to address complex queries is the first-order logical entailment.
For simplicity, our introduction \textbf{only} restricts to a finite entity set $\mathcal{E}$ and a finite binary relation set $\mathcal{R}$. No functions are mentioned.
Detailed presentation of first-order logic can be found in model theory literature~\citep{marker2002model}.

\subsection{First-order Logical Entailment}

\begin{definition}[First-order sentence]
The set of first-order formulae is the minimal set $F$ such that
\begin{compactenum}
    \item $p(s, o)\in F$, where $p$ is the relation, $s, o$ can be either an entity in $\mathcal{E}$ or a variable.
    \item If $s\in F$, then $\lnot s\in F$; If $s, t\in F$, then $s\land t \in F$ and $s\lor t \in F$.
    \item If\,$s(x)\in F$\,and\,$x$\,is\,a\,variable\,that\,is\,not\,qua- ntified, then $\forall x. s(x) \in F$ and $\exists x. s(x) \in F$.
\end{compactenum}
\end{definition}
We say a variable $x$ in a formula is quantified if there is a quantifier ($\exists x$ or $\forall x$). Otherwise, we say a variable is free. A sentence is a formula without free variables.

Next, we introduce first-order logical entailment, the \textit{central} concept of first-order reasoning. In general, the entailment is a process of \textit{verifying} the \textit{conlusion} (one sentence to verify) given the \textit{knowledge} (a set of given premises). \textbf{Notably, \textit{knowledge}, \textit{conclusion} and \textit{verification} are the three key components of the definition}. For first-order logical entailment, the \textit{knowledge} and \textit{conclusion} are restricted as first-order sentences, and the \textit{verification} process is subject to the first-order logical calculus. A set of FO sentences $\{p_i, i=1, ..., n\}$ entails a FO conclusion $s$ is denoted as:
\begin{align}
    \underbrace{\{p_i, i=1, ..., n\}}_{\text{ \textit{knowledge} or premises}} \models \underbrace{s}_{\text{\textit{conclusion}}}
\end{align}
After verification, its truth value suggests whether $s$ is the logical consequence of the premises.

\subsection{Knowledge Graph Query Answering}

Knowledge graph query answering is an important application of FO logical entailment. Its importance comes from both the importance of knowledge graphs and the queries in database systems. Here, we connect the task of KG query answering and FO entailment from the aspects of \textit{knowledge}, \textit{conclusion}, and \textit{verification}.

\noindent\textbf{\textit{Knowledge}.} Given an entity set $\mathcal{E}$ and relation set $\mathcal{R}$, a knowledge graph is defined as a set of triplets $\mathcal{G} = \{(h_i , r_i , t_i)\}$, where $h_i, t_i \in \mathcal{E}$ are entities and $r_i \in \mathcal{R}$ are relations. Each triple $(h, r, t)$ in KG can be regarded as the simplest form of first-order sentence $r(h, t)$. Thus, the knowledge graph $\mathcal{G}$ provided the premises.

\noindent\textbf{Answers to the query as multiple \textit{conclusions} to \textit{verify}.} Existing research for KG query answering usually focuses on the restricted families of FO queries, specifically, the Existential First-Order (EFO) queries. An EFO query is represented in Disjunctive Normal Form (DNF) as a disjunction of one or multiple conjunctions:
\begin{equation}
    q[V_?] = V_?.\exists V_1,V_2,...,V_n:c_1 \vee c_2 \vee ... \vee c_m
    \label{equa:efo}
\end{equation}
In equation \ref{equa:efo}, $V_?$ is the free variable, $V_1, V_2,..., V_n$ are existential variables, while $c_1, c_2,...,c_m$ are conjunctions of one-hop atomic formulas or their negation, defined as $c_i = \alpha_{i1} \wedge \alpha_{i2} \wedge ... \wedge \alpha_{ik_i}$ and $\alpha_{ij} = r(h, t)$ or $\neg r(h, t)$, where $r \in \mathcal{R}$, $h$ and $t$ are either constant entities $e \in \mathcal{E}$ or variables.

We can see that an EFO query is not a FO sentence because it contains a free variable to query. The goal of answering a query is to find all possible entities $a$ such that $\mathcal{G}\models q[a/V_?]$, where $q[a/V_?]$ is a FO sentence where the free variable $V_?$ is substituted by entity $a$. 
To summarize, the answer set 
\begin{equation}
    A(\mathcal{G},q) = \{a\in \mathcal{E}: \mathcal{G}\models q[a/V_?] = {\rm True}\}.
    \label{equa:answer_set}
\end{equation}

Next, we show that the key condition to be verified, i.e., $\mathcal{G}\models q[a/V_?] = {\rm True}$, relies on the knowledge graph $\mathcal{G}$. By FO logical calculus, evaluating $\mathcal{G}\models q[a/V_?]$ is equivalent to:

{\small
\begin{align}
    \bigvee_{V_1,..., V_n \in \mathcal{E}} \left[ \bigvee_{i=1}^m \left[ \left(\mathcal{G}\!\models\! \alpha_{i1}\right)_{V_?=a} \wedge ... \wedge \left(\mathcal{G}\!\models\! \alpha_{ik_i}\right)_{V_?=a} \right]\right].
\end{align}
}

We could see that whether $\mathcal{G}\models q[a/V_?]$ is True \textbf{essentially depends on} many times of FO logical entailment on atomic formulae $\mathcal{G}\models \alpha_{ij}$, where each of these depends on whether the atomic $r(h, t)$ associated to $\alpha_{ij}$ is contained in $\mathcal{G}$ (or not in $\mathcal{G}$ for the negation case).

\begin{figure*}[t]
\begin{center}
\includegraphics[clip,width=\linewidth]{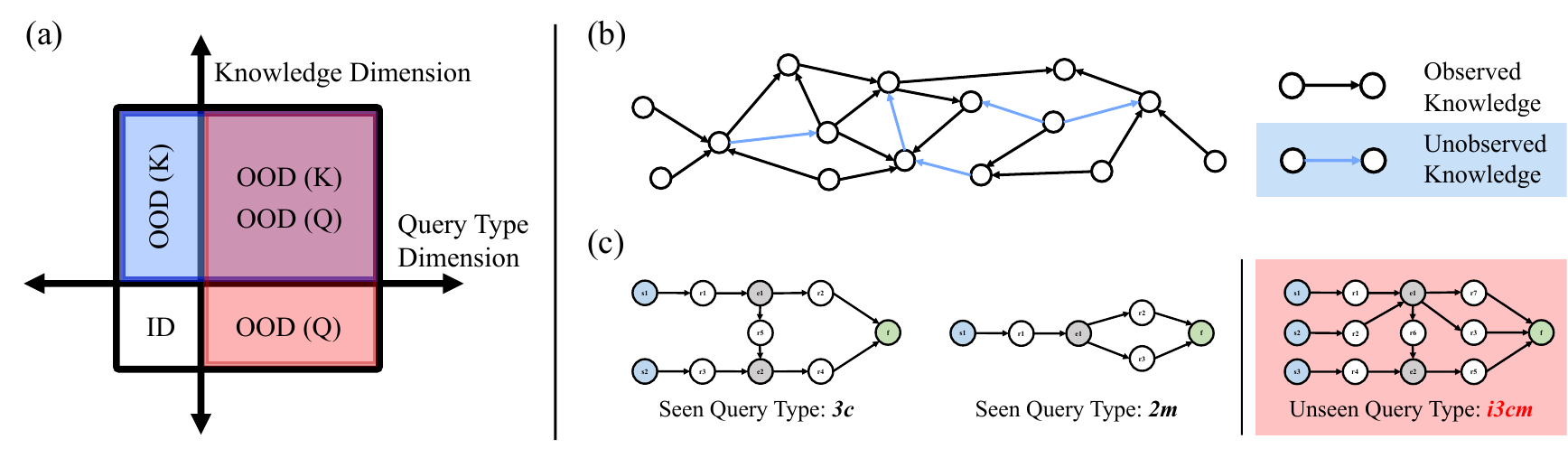}
\end{center}
\caption{An illustration of our two-fold \textit{Out-of-distribution Generalization} in knowledge and query type dimensions. Figure (a) explains the relation between OOD shifts in the two dimensions. Figure (b) and Figure (c) provide examples demonstrating the causes of OOD shifts in the two dimensions.
}
\label{fig:two_fold_gen}
\end{figure*}
\section{Distribution Shifts in KGQA}
\label{sec:shifts}
This section introduces the formulation of distribution shifts in KG query answering tasks. Such tasks are connected to first-order reasoning due to their close connection with logical entailment. By clearly presenting their connection with distribution shifts studied in statistical ML community~\citep{moreno2012unifying}, they are then able to measure the generalizable first-order reasoning capability of deep learning models, in particular transformers, with parameterized knowledge.

\noindent\textbf{Problem formulation.} A dataset for KG query answering is denoted as a set of samples $\{(x_i, y_i)\}$ drawn from the joint distribution $P(X, Y)$. Here, $X$ represents the random variable corresponding to an EFO query, with $x_i$ as individual samples, and $Y$ represents the random variable for the answer set, with $y_i$ as individual samples.

This paper studies transformers, where the input $X$ is tokenized and fed into the transformer architecture, and the output $Y$ is predicted with a classifier, where all entities in the KG are ranked by embedding similarities. In both the training and testing phases, the knowledge graph $\mathcal{G}$ is not explicitly accessed by the model. In this way, the necessary condition for transformers to correctly answer the queries is that the knowledge in KG is materialized into their parameters. To examine whether transformers possess the \textbf{generalizable} first-order reasoning capability, we consider two classic types of out-of-distribution shifts: concept shift and covariant shift. They are connected to unobserved knowledge and unseen query types.

\subsection{Two Types of Distribution Shifts}
Out-Of-Distribution (OOD) generalization is one of the key topics in machine learning research \cite{NIPS1991_ff4d5fbb,dataset_shift_in_ml}. It considers the shifts in the joint distribution $P(X, Y)$ from the training to the testing phases. Equation \ref{equa:ood_shift} decomposed the distribution shift by applying conditional probability formula, introducing two basic types of OOD shifts: \textit{concept shift} refers to the changes in $P(Y|X)$, while \textit{covariate shift} refers to the changes in $P(X)$.

{\small
\begin{align}
    P_{\rm train}(Y|X)&P_{\rm train}(X) \neq \underbrace{P_{\rm test}(Y|X)}_{\rm Concept} \underbrace{P_{\rm test}(X)}_{\rm Covariate}
    \label{equa:ood_shift}
\end{align}
}

\noindent As\,illustrated\,in\,Fig.\,\ref{fig:two_fold_gen},\,we\,studied\,distribution shifts in both knowledge and query type dimensions.

\paragraph{Concept Shift by Unobserved Knowledge}
In our KG query answering setting, we construct an \textit{observed} knowledge graph $\mathcal{G}_{o}$ as a subset of the full knowledge graph $\mathcal{G}$. In the training stage, the models only access the answers from the observed graph, denoted as $A(\mathcal{G}_o,q)$. While in the testing stage, they are evaluated on answers from the full graph, denoted as $A(\mathcal{G},q)$. This setup introduces a concept shift in the conditional distribution $P(Y|X)$ between training and testing because the change of knowledge causes different results of entailment, see Equation~\eqref{equa:answer_set}. 
Therefore, the set difference $\mathcal{A}_{ood} = A(\mathcal{G},q) - A(\mathcal{G}_o,q)$ contains the entities that can only be derived when the model generalizes from knowledge $\mathcal{G}_o$ to $\mathcal{G}$.
By measuring the performances of models on $\mathcal{A}_{ood}$, we are able to access how a model generalizes under the concept shift caused by unobserved knowledge.
This is also termed \textit{knowledge generalization} or \textit{knowledge inference} in the literature \cite{sun2021faithful}.
\setlength{\tabcolsep}{4.4pt}
\begin{table}[t]
\scriptsize
\begin{tabular}{lcccccccc}
\toprule
\multicolumn{1}{c}{\multirow{2}{*}{Dataset}} & \multicolumn{4}{c}{Operations} & \multicolumn{3}{c}{Features} & \multirow{2}{*}{Unseen Types} \\
  & pro. & int. & uni. & neg. & exi. & mul. & cyc. &  \\ \midrule
GQE  & \checkmark & \checkmark & \multicolumn{1}{l}{} & \multicolumn{1}{l}{} &  &  & & 0 \\
Q2B  & \checkmark & \checkmark & \checkmark & \multicolumn{1}{l}{} &  &  & & 4 \\
BetaE & \checkmark & \checkmark & \checkmark & \checkmark &  &  & & 4 \\
FIT & \checkmark & \checkmark & \checkmark & \checkmark & \multicolumn{1}{c}{\checkmark} & \multicolumn{1}{c}{\checkmark} & \multicolumn{1}{c}{\checkmark} & 10 \\
SQE & \checkmark & \checkmark & \checkmark & \checkmark &  &  & &  29 \\ \midrule
Ours  & \checkmark & \checkmark & \checkmark & \checkmark & \multicolumn{1}{c}{\checkmark} & \multicolumn{1}{c}{\checkmark} & \multicolumn{1}{c}{\checkmark}  &\textbf{32} \\ \bottomrule
\end{tabular}
\caption{Comparison between different benchmarks on their supported logical operations, query features, and the number of unseen query types.}
\label{tab:benchmark_comparison}
\end{table}
\paragraph{Covariate Shift by Unseen Query Types}
We also examine the capability of query answering models to generalize and maintain performance on query types that are unseen in the training stage. In our testing data, we include new query types that are combinatorially crafted from the training query types, introducing a covariate shift in the distribution of $P(X)$. The models' performances on these unseen types are measured to evaluate their generalizability to covariate shifts in the query type dimension. This is also termed \textit{combinatorial generalization} or \textit{compositional generalization} in the literature \cite{wang2021benchmarking, yin2023textefokcqa}.

\section{Evaluation Benchmark}
\label{sec:benchmark}
To better investigate the performance in generalizable FO reasoning, we build a new benchmark of KG query answering for our experiment. We first introduce the dataset construction process and evaluation settings, and then present our results of transformers in comparison to previous methods.
\setlength{\tabcolsep}{7.7pt}
\begin{table*}[t]
\centering
\scriptsize
\begin{tabular}{clccccccc}
\toprule
\multirow{2}{*}{Dataset} & \multicolumn{1}{c}{\multirow{2}{*}{Model}} & \multirow{2}{*}{Type} & \multicolumn{2}{c}{ID (Q)} & \multicolumn{2}{c}{OOD (Q)} & \multicolumn{2}{c}{All Queries} \\\cmidrule{4-9}
 & \multicolumn{1}{c}{} &  & ID (K) & OOD (K) & ID (K) & OOD (K) & ID (K) & OOD (K) \\ \midrule
\multirow{9}{*}{FB15k} & BetaE \cite{ren2020beta}& Probabilistic & 26.9	& 18.5 & 22.4 & 13.5 & 24.3 & 15.6 \\
& ConE \cite{zhang2021cone}& Geometric & 35.5  & 22.0  & 27.2 & 15.6 & 30.7  & 18.3  \\
 & CQD \cite{arakelyan2021complex}& Fuzzy Logic & 33.1  & 20.7 & 21.5  & 11.2  & 26.4  & 15.2  \\
 & LMPNN \cite{wang2023logical}& GNN & 32.5  & 21.0 & 24.3  & 13.9  & 27.7  & 16.9  \\
 & SQE-LSTM \cite{DBLP:journals/corr/abs-2302-13114}& RNN & 39.9  & 26.3  & 31.5 & {\ul 18.5}  & 35.0  & 21.8 \\ \cmidrule{2-9} 
 & Trans.+Absolute PE & Transformer & 46.9  & 31.9 & 21.8  & 13.2  & 32.3  & 21.0  \\
 & Trans.+Disentangled PE & Transformer & \textbf{51.7} & \textbf{33.3} & 23.7 & 13.6  & 35.4  & 21.8  \\
  & Trans.+Rotary PE & Transformer & {\ul 50.1}  & {\ul 32.7}  & \textbf{34.6}  & 20.8  & \textbf{41.1}  & {\ul 25.8}  \\
 & Trans.+Relative PE & Transformer & 48.1  & 32.3  & {\ul 35.4}  & \textbf{21.5}  & {\ul 40.7}  & \textbf{26.0}  \\
 \midrule
\multirow{9}{*}{FB15k-237} & BetaE \cite{ren2020beta}& Probabilistic & 27.4 & 16.7 & 23.3 & 13.1 & 25.0 & 14.6 \\
& ConE \cite{zhang2021cone}& Geometric & 34.9 & 18.5 & 27.3 & 14.2 & 30.5 & 16.0 \\
 & CQD \cite{arakelyan2021complex}& Fuzzy Logic & 32.8 & 14.1 & 20.2 & 8.6 & 25.5 & 10.9 \\
 & LMPNN \cite{wang2023logical}& GNN & 26.6 & 16.4 & 22.3 & 12.4 & 24.1 & 14.1 \\
 & SQE-LSTM \cite{DBLP:journals/corr/abs-2302-13114}& RNN & 45.8 & 17.9 & {\ul 32.3} & {\ul 14.6} & 37.9 & {\ul 16.0} \\ \cmidrule{2-9} 
 & Trans.+Absolute PE & Transformer & 54.4 & {\ul 19.9} & 20.8 & 9.0 & 34.9 & 13.6 \\
 & Trans.+Disentangled PE & Transformer & \textbf{59.0} & \textbf{20.0} & 21.3 & 8.7 & 37.0 & 13.4 \\
  & Trans.+Rotary PE & Transformer & 54.3 & 19.8 & 36.7 & 15.4 & {\ul 44.1} & 17.2 \\
 & Trans.+Relative PE & Transformer & {\ul 54.8} & \textbf{20.0} & \textbf{37.7} & \textbf{15.8} & \textbf{44.8} & \textbf{17.6} \\
 \midrule
\multirow{9}{*}{NELL995} & BetaE \cite{ren2020beta}& Probabilistic & 40.9 & 16.7 & 32.3 & 11.1 & 35.9 & 13.4 \\
& ConE \cite{zhang2021cone}& Geometric & 46.1 & 17.9 & 38.3 & {\ul 12.8} & 41.5 & 14.9 \\
 & CQD \cite{arakelyan2021complex}& Fuzzy Logic & 42.9 & 14.7 & 28.9 & 8.3 & 34.8 & 11.0 \\
 & LMPNN \cite{wang2023logical}& GNN & 42.8 & 17.1 & 32.1 & 11.6 & 36.6 & 13.9 \\
 & SQE-LSTM \cite{DBLP:journals/corr/abs-2302-13114}& RNN & 65.1 & 17.8 & 48.3 & 12.5 & 55.3 & {\ul 14.7} \\ \cmidrule{2-9} 
 & Trans.+Absolute PE & Transformer & 68.8 & 18.9 & 25.8 & 8.3 & 43.8 & 12.7 \\
 & Trans.+Disentangled PE & Transformer & \textbf{73.6} & \textbf{19.3} & 29.7 & 8.7 & 48.1 & 13.1 \\
  & Trans.+Rotary PE & Transformer & {\ul 69.4} & {\ul 19.3} & \textbf{50.6} & 13.2 & \textbf{58.4} & \textbf{15.7} \\ 
 & Trans.+Relative PE & Transformer & 68.0 & 18.7 & {\ul 49.8} & \textbf{13.4} & {\ul 57.4} & {\ul 15.6} \\
\bottomrule
\end{tabular}
\caption{Experimental result (in MRR\%) of Transformers in comparison with previous state-of-the-art methods with different backbones in our full benchmark. Definitions of evaluation metrics are provided in Sec. \ref{sec:metrics}.}
\label{tab:benchmark_result}
\end{table*}
\subsection{Dataset Construction}
Table \ref{tab:benchmark_comparison} compares the feature and statistics of query types selected in our benchmark with previous benchmarks in KG query answering, namely GQE dataset \cite{hamilton2019embedding}, Q2B dataset \cite{ren2020query2box}, BetaE dataset \cite{ren2020beta}, FIT dataset \cite{yin2023rethinking}, and SQE dataset \cite{DBLP:journals/corr/abs-2302-13114}. We included the entire set of set operations and query features in our selection of 23 seen query types. Moreover, we \textbf{compositionally crafted 32 unseen query types} that are only used in the testing stage to examine models' OOD generalizability in the query type dimension. For instance\footnote{Queries are represented in EFO syntax \cite{yin2023rethinking}.}, suppose a query answering model is trained on query type \textit{2in} [\texttt{(r1(s1,f))\&(!(r2(s2,f)))}] and \textit{2p} [\texttt{(r1(s1,e1))\&(r2(e1,f))}], we expect it can also handle the unseen query type \textit{in2p} [\texttt{(r1(s1,e1))\&(!(r2(s2,e1)))\&(r3(e1,e2))\&\\(r4(e2,f))}], which is compositionally designed based on the previous two. All queries are sampled from the following three knowledge graphs: FB15k \cite{Bollacker2008FreebaseAC,NIPS2013_1cecc7a7}, FB15k-237 \cite{toutanova-chen-2015-observed}, and NELL995 \cite{10.5555/2898607.2898816}. The details of all query types and their statistics in our benchmark are provided in Appendix \ref{app:benchmark_details}. We also provide query graphs for graph-augmented methods \cite{Liu_2022,xu-etal-2023-query2triple} in our experiments, with detailed definitions in Appendix \ref{app:query_graph}.

\subsection{Evaluation Settings}
\label{sec:metrics}
In accordance with the two types of distribution shifts discussed above, we evaluate the performance of query answering models across the two corresponding dimensions using the mean reciprocal rank (MRR) metric. In the knowledge dimension, we employ the notation ID (K) and OOD (K) to represent the performance on the answer set $\mathcal{A}_{id} = A(\mathcal{G}_o,q)$ and $\mathcal{A}_{ood} = A(\mathcal{G},q) - A(\mathcal{G}_o,q)$, respectively. Concurrently, in the query type dimension, we utilize the notation ID (Q) and OOD (Q) to represent the performance on seen query types and unseen query types, respectively. Higher scores in ID (K) and ID (Q) reflect the models' effectiveness in performing logical entailment on in-distribution data, while higher scores in OOD (K) and OOD (Q) demonstrated better generalizability of the models. It is also noteworthy to mention that settings in ID (K) and OOD (K) denote the concept shifts, while those in ID (Q) and OOD (Q) capture the covariate shift. For details regarding the metric calculation, please refer to Appendix \ref{app:evaluation_details}.
\begin{figure*}[t]
\begin{center}
\includegraphics[clip,width=\linewidth]{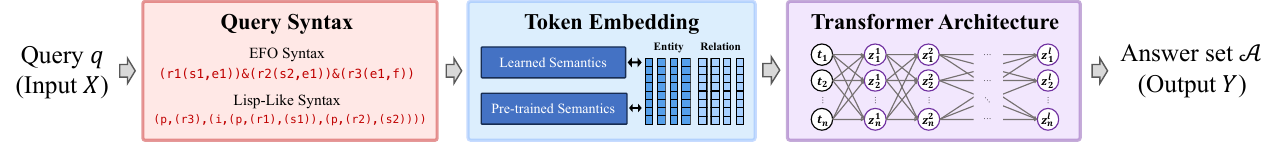}
\end{center}
\caption{An illustration of three core stages in the pipeline of KG query answering with transformers.}
\label{fig:pipeline}
\end{figure*}
\subsection{General Benchmarking Results}
We evaluate transformer encoders\footnote{Decoder-only and Encoder-Decoder yield comparable performance as illustrated in Appendix \ref{app:transformer_arch}.} against five existing methods for KG query answering, with details provided in Appendix \ref{app:baseline}. Meanwhile, we compare four positional encoding settings in transformers: \textbf{Absolute PE} is the default sinusoidal positional encoding introduced in the original transformer \cite{vaswani2023attention}. \textbf{Relative PE} \cite{shaw2018selfattention} applied a learnable matrix based on the relative position of tokens. \textbf{Disentangled PE} \cite{he2021deberta} uses separate vectors to encode the content and position of each token, decoupling the contextual and positional information during self-attention. \textbf{Rotary PE} \cite{su2023roformerenhancedtransformerrotary} encodes position information by rotating query/key vectors in the complex space, capturing relative positions through rotation angles.

The experimental results are presented in Table \ref{tab:benchmark_result}. Transformers marginally outperform all baselines in both ID (K) and OOD (K) in seen query types. Both APE and DPE transformers have poor OOD generalizability in the query type dimension. In contrast, RPE and RoPE transformers demonstrates much stronger generalizability, as it can better capture structural information from the distances between tokens and flexibly handle query sequences of unseen lengths. Overall, RPE transformers achieve the best performance and OOD generalizability on first-order logical entailment.

\section{Transformer Architectures and First-order Reasoning Capability}
\label{sec:experiments_and_analyses}

In this section, we further investigate the dependencies between transformers' performance and multiple design choices throughout the modeling process. Figure~\ref{fig:pipeline} depicts the three-stage pipeline for deriving the answer set $\mathcal{A}$ (output $Y$) of query $q$ (input $X$) using transformers. We present the impact of query syntax in Section \ref{sec:query_syntax}, token embeddings in Section \ref{sec:token_embeddings}, and transformer architecture in Section \ref{sec:transformer_architecture}. We summarize the findings in Section \ref{sec:sum_findings}.

\subsection{Study on Query Syntax}
\label{sec:query_syntax}
Logical queries with the same semantics can be represented in different syntaxes with different formal languages. To investigate the impact of difference in formal languages, we conducted experiments on two formal languages: \textbf{Lisp-like Syntax} \citep{wang2021benchmarking} represents query with fully parenthesized nested formula; \textbf{EFO Syntax} \cite{yin2023rethinking} represents query with all one-hop atomics connected with conjunctions and disjunctions in parallel. As the Lisp-like syntax has a limited representation scope, we used a subset of our benchmark that includes 13 seen query types and 12 unseen query types that are supported by both languages. The experimental results on the impact of formal language are presented in Table \ref{tab:formal-language}. In APE Transformers, there is no difference between the results of the two formal languages. However, with RPE, the EFO syntax achieves much better generalizability in the query type domain. This phenomenon can be explained by the differences in the structural features of the two languages: with a parallel structure, the distances (i.e., relative positions) between tokens with the same logical relationship are more consistent than those in a nested structure. As a result, RPE can better learn the logical relationship between tokens and better generalize to unseen query types.
\setlength{\tabcolsep}{3.8pt}

\begin{table}[t]

\centering
\scriptsize
\begin{tabular}{cccccc}
\toprule
\multirow{2}{*}{Model} & \multicolumn{1}{c}{\multirow{2}{*}{Syntax}} & \multicolumn{2}{c}{ID (Q)} & \multicolumn{2}{c}{OOD (Q)} \\\cmidrule{3-6}
 & \multicolumn{1}{c}{} & \multicolumn{1}{c}{ID (K)} & \multicolumn{1}{c}{OOD (K)} & \multicolumn{1}{c}{ID (K)} & \multicolumn{1}{c}{OOD (K)}  \\ \midrule
\multirow{2}{*}{Trans.+Absolute PE} & Lisp-Like & 57.1 & \textbf{20.7} & 10.0 & 4.9 \\
 & EFO & \textbf{58.1} & \textbf{20.7} & \textbf{10.4} & \textbf{5.1}  \\ \midrule
\multirow{2}{*}{Trans.+Relative PE} & Lisp-Like & \textbf{58.3} & \textbf{20.9} & 22.1 & 9.8  \\
 & EFO & 58.1 & \textbf{20.9} & \textbf{35.4} & \textbf{14.3}\\ \bottomrule

\end{tabular}
\caption{Experimental results of Transformers in FB15k-237 with different formal languages. Details of the dataset fragment are provided in Appendix \ref{app:benchmark_details}.}
\label{tab:formal-language}
\end{table}

\setlength{\tabcolsep}{7pt}

\begin{table}[t]
\centering
\scriptsize
\begin{tabular}{ccccc}
\toprule
\multirow{2}{*}{Model} & \multicolumn{2}{c}{Before Reversion} & \multicolumn{2}{c}{After Reversion} \\ \cmidrule(l){2-5} 
 & ID(K) & OOD(K) & ID(K) & OOD(K) \\ \midrule
Trans.+Absolute PE & 54.1 & 18.3 & 27.8 & 9.8 \\
Trans.+Relative PE & 54.3 & 17.0 & 54.5 & 16.8 \\ \bottomrule
\end{tabular}
\caption{Experiment results of Transformers on five query types before and after reversing permutation.}
\label{tab:perm-inv}

\end{table}

Furthermore, even for the same query in EFO Syntax, there can be various permutations of one-hop atomics, with the total number of potential permutations growing factorially with the length of the query. Therefore, it is crucial for transformers to robustly handle queries with different permutations. To evaluate this property, we selected five query types (\textit{2p}, \textit{3p}, \textit{ip}, \textit{pi}, \textit{2in}) and reversed their permutations while maintaining the same semantics in the testing data (see Appendix \ref{app:reversion}). The experimental results on query permutation are presented in Table \ref{tab:perm-inv}. We observed a significant performance drop after reversing the query in APE transformers. However, for RPE transformers, the performance before and after reversing the permutation remains consistent, demonstrating their robustness in handling queries with modified permutations.

\subsection{Study on Token Embeddings}
\label{sec:token_embeddings}
Some methods in KG query answering \cite{arakelyan2021complex,wang2023logical,xu-etal-2023-query2triple} utilized pre-trained KG embeddings from link predictors in their query encoding process, while other methods \cite{ren2020beta,zhang2021cone,DBLP:journals/corr/abs-2302-13114} randomly initialize the embedding matrix and jointly learn the KG embedding and model parameters from the training process. It's still unclear that to what extent can pre-trained KG semantics can assist the performance of query answering models. To rigorously assess the impact of pre-trained semantics in token embeddings, we have selected three widely recognized pre-trained KG embeddings: \textbf{TransE} \cite{NIPS2013_1cecc7a7} models relationships by interpreting them as translations within the embedding space, \textbf{DistMult} \cite{yang2015embedding} employs a bilinear model wherein entities are represented as vectors and relations as diagonal matrices, \textbf{ComplEx} \cite{trouillon2016complex} extends this framework into a complex vector space, allowing for more nuanced inter-entity interactions. In our experiment, all three KG embeddings are frozen during the training stage, in comparison to a baseline with randomly initialized and jointly learned embeddings.

The experimental results are presented in Table \ref{tab:kg-embedding}. Two stronger KGEs, ComplEx and DistMult, can effectively improve the performance of transformer. In contrast, the performance of TransE consistently lags behind the baseline across all three datasets. This discrepancy can be attributed to the baseline model's embedding matrix learning process, which can be viewed as implicitly training a link predictor using a transformer architecture (e.g., KG-BERT \cite{yao2019kgbert}). This modern approach may naturally outperform earlier models like TransE.
\setlength{\tabcolsep}{5.5pt}

\begin{table}[t]
\centering
\scriptsize
\begin{tabular}{cccccc}
\toprule
\multirow{2}{*}{Dataset} & \multicolumn{1}{c}{\multirow{2}{*}{KGE}} & \multicolumn{2}{c}{ID (Q)} & \multicolumn{2}{c}{OOD (Q)} \\\cmidrule{3-6}
 & \multicolumn{1}{c}{} & ID (K) & OOD (K) & ID (K) & OOD (K) \\ \midrule
\multirow{4}{*}{FB15k} & Random & 48.1 & 32.3 & 35.4 & 21.5 \\
 & TransE & 44.3 & 30.3 & 34.2 & 20.8 \\
 & DistMult & \textbf{49.7} & \textbf{32.6} & \textbf{37.5} & \textbf{21.8} \\
 & ComplEx & {\ul 49.5} & {\ul 32.4} & {\ul 36.4} & {\ul 21.6} \\ \midrule
\multirow{4}{*}{FB15k-237} & Random & 54.8 & 20.0 & 37.7 & 15.8 \\
 & TransE & 54.5 & 20.0 & 35.0 & 14.9 \\
 & DistMult & {\ul 57.2} & \textbf{20.4} & {\ul 39.8} & \textbf{16.1} \\
 & ComplEx & \textbf{58.6} & \textbf{20.4} & \textbf{40.4} & {\ul 16.0} \\ \midrule
\multirow{4}{*}{NELL995} & Random & 68.0 & 18.7 & 49.8 & \textbf{13.4} \\
 & TransE & 68.5 & 18.6 & 48.7 & 12.5 \\
 & DistMult & \textbf{72.2} & \textbf{19.9} & \textbf{51.2} & 12.9 \\
 & ComplEx & \textbf{72.2} & {\ul 19.8} & {\ul 49.9} & {\ul 13.1} \\ \bottomrule
\end{tabular}
\caption{Experimental results of Transformer + Relative PE with different pre-trained KGEs.}
\label{tab:kg-embedding}
\end{table}
\subsection{Study on Transformer Architecture}

\label{sec:transformer_architecture}
Some existing literature \cite{Kotnis_Lawrence_Niepert_2021,Liu_2022,xu-etal-2023-query2triple} has explored the use and design of transformers in KG query answering, in which several inductive biases in transformer architecture were proposed. However, as all these methods focused on different aspects and were evaluated on different benchmarks, a clearer investigation is needed. We implemented two designs from previous works on top of transformers: \textbf{Adjacency Matrices Masking}: an inductive bias proposed in kgTransformer \cite{Liu_2022}, which aims to enhance reasoning performance by limiting the range of self-attention to one-hop neighboring nodes in the query graph. \textbf{Directed Distance Encoding}: another inductive bias, introduced in Query2Triple \cite{xu-etal-2023-query2triple}, that facilitates information aggregation by injecting a directed distance message into the self-attention mechanism. The experimental results on existing transformer designs are presented in Table \ref{tab:exist-biases}. Both inductive biases effectively improve generalizability in the query type dimension when using APE. In contrast, under RPE, these improvements are not observed, and the use of adjacency matrices masking even results in a significant decrease in performance. This decrease is attributed to the masking in self-attention, which restricts message passing to only nodes within one hop, thereby limiting the model's ability to perform complex logical reasoning. Regarding directed distance encoding, the information about directed distances is mostly contained in the relative positions of the linearized sequence, thus the improvements seen with original APE are not sustained.

\setlength{\tabcolsep}{4pt}

\begin{table}[t]
\centering
\scriptsize
\begin{tabular}{lcccc}
\toprule
 \multicolumn{1}{c}{\multirow{2}{*}{Model}} & \multicolumn{2}{c}{ID (Q)} & \multicolumn{2}{c}{OOD (Q)} \\\cmidrule{2-5}
& \multicolumn{1}{c}{ID (K)} & \multicolumn{1}{c}{OOD (K)} & \multicolumn{1}{c}{ID (K)} & \multicolumn{1}{c}{OOD (K)}  \\ \midrule
  Trans.+Absolute PE (baseline) & {\ul 54.4} & {\ul 19.9} & 20.8 & 9.0\\
  +Adjacency Matrices Masking & 46.5 & 19.6 & {\ul 22.1} & \textbf{11.2} \\
  +Directed Distance Encoding & \textbf{55.0} & \textbf{20.1} & \textbf{22.7} & {\ul 9.8} \\ \midrule
 Trans.+Relative PE (baseline) & \textbf{54.8} & {\ul 20.0} & \textbf{37.7} & {\ul 15.8} \\
 +Adjacency Matrices Masking & 51.3 & \textbf{20.1} & 28.3 & 13.0\\
 +Directed Distance Encoding & 54.1 & {\ul 20.0} & {\ul 37.5} & \textbf{15.9}  \\ \bottomrule
\end{tabular}
\caption{Experimental result of Transformer + Absolute/Relative PE in FB15k-237 with existing inductive biases. Considering its triviality and compatibility with our dataset, the inductive bias from BiQE is not included in our experiment.}

\label{tab:exist-biases}
\end{table}
\subsection{Summary of Empirical Findings}
\label{sec:sum_findings}
In terms of query syntax, we discovered that \textbf{employing formal language with parallel structure can significantly enhance OOD generalizability} in the query type dimension for RPE transformers. Moreover, we observed that \textbf{transformers with relative PE can robustly handle queries with different permutations}. Regarding transformer architecture, we found that the inductive biases proposed in previous approaches are \textbf{only effective under APE, but fail to improve upon RPE transformers}, which we have shown to be the preferred setting. This mismatch between transformer design and positional encoding settings motivated us to design a methodology that effectively boosts the performance and generalizability of RPE transformers in first-order logical entailment.
\section{Transformer Encoder with Guided Attention}
\label{sec:tega}
One of the main challenges in KG query answering is to perform multi-hop logical reasoning following the one-hop atomics and the logical operations between them. In traditional reasoning tasks, such as language-based reasoning, transformer models perform multi-hop reasoning by procedurally forming ``reasoning trees'' within their attention mechanisms \cite{hou2023mechanistic,murty-etal-2023-grokking,wang2023label}. While in our task, it is challenging for transformers to capture the pattern of reasoning trees behind the queries, as the semantics of logical operators are neither explicitly modeled nor pre-trained, but implicitly learned from limited training queries. To navigate this conundrum, we present \textbf{TEGA} (\textbf{T}ransformer \textbf{E}ncoder with \textbf{G}uided \textbf{A}ttention), a modeling methodology that boosts logical reasoning performance by providing logic-aware guidance within the self-attention mechanism of transformers. The following sections will introduce the two inductive biases in TEGA and their effectiveness with experimental results.

\subsection{Logic-aware Relative Positional Encoding} 
Our analysis so far shows that RPE is the best positional encoding for transformers in our task for its superior performance and generalizability in both knowledge and query type dimensions. However, it still has limited expressiveness in the logical relationship between tokens, as the tokens with the same relative distance can have different logical relationships among queries. To fill this gap, we propose LogiRPE, a logic-aware mechanism to enhance self-attention. Specifically, for a sequence input sequence embeddings $x_i\in \mathbb{R}^{d}$, $i=1,..., n$, the enhanced self-attention is computed by

{\small
\begin{equation}
    z_i = \sum_{j=1}^n \alpha_{ij}(x_jW^V + \beta_{ij}^V), \quad \alpha_{ij} = \frac{\exp{e_{ij}}}{\sum^n_{k=1}\exp{e_{ik}}}
\end{equation}
\begin{equation}
\quad e_{ij} = \frac{x_iW^Q(x_jW^K + \beta_{ij}^K)^T}{\sqrt{d}}.
\end{equation}
}
where $W^Q, W^K, W^V\in \mathbb{R}^{d\times d}$ are the weight matrices for query, key, and value while $\beta_{ij}^V, \beta_{ij}^K \in \mathbb{R}^{d}$ are two \textit{logical bias} terms introduced by LogiRPE to capture the missing logical relations.

The key feature of \textit{logical bias} terms is that they should be able to discriminate different types of tokens. Let the type of the $i$-th token be $t_i$ from a finite type set of size $t$. In our study, we consider six types of tokens: parenthesis\texttt{[(,)]}, entity\texttt{[s,e,f]}, relation\texttt{[r]}, conjunction\texttt{[$\land$]}, disjunction\texttt{[$\lor$]}, and negation\texttt{[!]}. Then the bias vector $\beta_{ij}^{l}$ ($l$ is either $K$ or $V$) is indexed from an trainable embedding bank $\omega^{l}\in\mathbb{R}^{t\times t\times n\times d}$ (consider the first three indices).
\begin{align}
    \beta_{ij}^{l} = \omega^{l}[t_i, t_j, |i-j|], \quad l\in \{K, V\}.
\end{align}
LogiRPE may also be applicable to broader tasks where tokens with logical semantics can be labeled.
\subsection{Free-variable Pooling}
Most existing transformer-based approaches \cite{DBLP:journals/corr/abs-2302-13114,xu-etal-2023-query2triple} utilize the last layer hidden states of the first token(s) as the final query encoding, akin to the use of the $[CLS]$ token in the original BERT paper \cite{devlin2019bert}. However, we contend that the reasoning capabilities inherent in self-attention mechanisms should effectively aggregate the query information into free variables that semantically represent the answer sets of queries. Consequently, we adopt Free-Variable Pooling, wherein the final query encoding is derived through max pooling across the last-layer hidden states of these free variables.
\subsection{Result and Ablation}
The experimental results of TEGA and Transformer baselines are presented in Table \ref{tab:tega-result}. In all three datasets, TEGA substantially improved the performance and generalizability of knowledge and query type dimensions over baselines. According to our ablation study results in Table \ref{tab:tega-ablation}, both inductive biases proposed in TEGA are effective on their own. At the same time, a stronger improvement in entailment performance can be achieved when applied together. Moreover, we provide an example of attention visualization in Appendix \ref{app:attn_vis}, demonstrating the self-evident effect of logic-aware attention guidance within our TEGA model.

\setlength{\tabcolsep}{3.2pt}

\begin{table}[]
\centering
\scriptsize
\begin{tabular}{clcccc}
\toprule
\multirow{2}{*}{Dataset} & \multicolumn{1}{c}{\multirow{2}{*}{Model}} & \multicolumn{2}{c}{ID (Q)} & \multicolumn{2}{c}{OOD (Q)}  \\\cmidrule{3-6}
 & \multicolumn{1}{c}{} & ID (K) & OOD (K) & ID (K) & OOD (K) \\ \midrule
\multirow{3}{*}{FB15k} & Trans.+Absolute PE & 46.9  & 31.9 & 21.8  & 13.2 \\
& Trans.+Relative PE & 48.1 & 32.3 & 35.4 & 21.5 \\
 & TEGA (ours)& \textbf{55.0} & \textbf{34.1} & \textbf{38.8} & \textbf{22.4}  \\ \midrule
\multirow{3}{*}{FB15k-237}
 & Trans.+Absolute PE & 54.4 & 19.9 & 20.8 & 9.0  \\
 & Trans.+Relative PE & 54.8 & 20.0 & 37.7 & 15.8  \\
 & TEGA (ours)& \textbf{64.3} & \textbf{20.1} & \textbf{42.6} & \textbf{16.0}  \\ \midrule
\multirow{3}{*}{NELL995}  & Trans.+Absolute PE & 68.8 & 18.9 & 25.8 & 8.3 \\
& Trans.+Relative PE & 68.0 & 18.7 & 49.8 & 13.4 \\
 & TEGA (ours)& \textbf{75.0} & \textbf{19.2} & \textbf{56.6} & \textbf{13.7} \\ \bottomrule
\end{tabular}
\caption{Experimental results of TEGA comparing to baselines in FB15k, FB15k-237 and NELL995.}
\label{tab:tega-result}
\end{table}
\setlength{\tabcolsep}{3.8pt}

\begin{table}[t]
\centering
\scriptsize
\begin{tabular}{lcccccc}
\toprule
\multicolumn{1}{c}{\multirow{2}{*}{Model}} & \multicolumn{2}{c}{ID (Q)} & \multicolumn{2}{c}{OOD (Q)} \\\cmidrule{2-5}
\multicolumn{1}{c}{} & ID (K) & OOD (K) & ID (K) & OOD (K) \\ \midrule
TEGA (ours) & \textbf{55.0} & \textbf{34.1} & \textbf{38.8} & \textbf{22.4} \\ \midrule
w/o LogiRPE & {\ul 51.9} & {\ul 33.2} & {\ul 37.0} & 21.6 \\
w/o Free-Variable Pooling & 50.9 & 33.1 & 36.5 & {\ul 22.2}  \\ \midrule
Trans.+Relative PE (baseline) & 48.1 & 32.3 & 35.4 & 21.5  \\ \bottomrule
\end{tabular}
\caption{Ablation study on two inductive biases introduced in TEGA on FB15k.}
\label{tab:tega-ablation}
\end{table}
\section{Related Work}
\paragraph{Transformers in Logical Reasoning}
Transformers have exhibited remarkable performance in various forms of logical reasoning. In natural language inference (NLI) tasks \cite{bowman2015large,williams2018broadcoverage}, transformers analyze the logical relationship between a premise and a hypothesis, employing deductive reasoning over statements. Furthermore, transformers have been employed for inductive reasoning within rule-based systems \cite{clark2020transformers}. Additionally, transformers have been applied to perform abductive reasoning in both natural language \cite{bhagavatula2020abductive} and formal language \cite{bai2024advancing,gao2025controllablelogicalhypothesisgeneration} settings. Meanwhile, various techniques have been proposed to enhance logical reasoning in language understanding \cite{chen2023learning,pan2023logiclm}. Recent studies show that large language models (LLMs) exhibit enhanced logical reasoning abilities \cite{parmar2024logicbenchsystematicevaluationlogical,li2025patternsprinciplesfragilityinductive,fan2025legalruleinductiongeneralizable}, even without intermediate reasoning steps \cite{zheng2025cursecotlimitationschainofthought}. Logical entailment also plays a key role in hypothesis verification of LLMs, in areas such as commonsense reasoning \cite{zhao2024uncommonsensereasoningabductivereasoning}, analogical reasoning \cite{zheng2025logidynamicsunravelingdynamicslogical}, and even scientific discovery \cite{takagi2023autonomoushypothesisverificationlanguage, zheng2025automationautonomysurveylarge}.

\paragraph{KGQA with Parameterized Knowledge}
Prior to this work, extensive research has been conducted on KGQA with parameterized knowledge. Regarding modeling approaches, iterative neural query encoders \cite{ren2020beta, chen2022fuzzy, tsang2025transformerscomplexqueryanswering} design representations for entity sets and execute logical operators iteratively, following the computational graph. In addition, neural-symbolic methods \cite{zhu2022neuralsymbolic,bai2023answering,yin2023rethinking} incorporate link predictors and search over the symbolic space. In terms of the knowledge domain, existing benchmarks \cite{ren2020query2box,ren2020beta,yin2023textefokcqa} primarily utilize knowledge graphs containing general factual knowledge, while some studies have focused on commonsense knowledge \cite{fang2024complex} and eventuality knowledge \cite{bai2023complex}. Such queries can be extended to natural language settings with template or LLM-based approaches \cite{zheng2024clrfactevaluatingcomplexlogical,zheng2025knowshiftqarobustragsystems, zong2025comparisonqaevaluatingfactualityrobustness, bai2025autoschemakgautonomousknowledgegraph}. The setting of KGQA with parameterized knowledge has also been extended to the concept of \textit{Neural Graph Databases} \cite{ren2023neuralgraphreasoningcomplex,bai2025challengesagenticneuralgraph}.
\section{Conclusion}
This paper investigates the generalizable first-order logical reasoning capabilities of transformers when knowledge is parameterized within model weights. By establishing connections between distribution shifts and logical entailment, we develop a principled framework for evaluating transformer performance on knowledge graph query answering tasks. Our comprehensive benchmark reveals that transformers can effectively perform first-order logical entailment, outperforming methods specifically designed for this task. Through systematic analysis of the entire modeling pipeline, we identify critical design principles that enhance reasoning capabilities. Notably, our findings reveal a mismatch between existing inductive biases and optimal positional encoding strategies, motivating the development of TEGA. This architecture introduces logic-aware self-attention mechanisms that substantially improve both performance and generalizability. Our work demonstrates that careful architectural choices, informed by the logical structure of reasoning tasks, can significantly enhance transformers' logical entailment ability.
\section*{Limitations}
The scope of complex logical queries was limited to EFO queries with a single free variable, with queries containing multiple free variables reserved for future investigation. While LogiRPE incorporates fine-grained logic awareness, it remains dependent on explicit labeling. Future research should explore more generalizable approaches to logic-aware labeling that can accommodate previously unseen logical operators. With respect to knowledge graph scale, the largest knowledge graph employed in our experiments was FB15k, which is considered medium-sized within the knowledge graph literature. Consequently, the scalability of the proposed approaches to large-scale knowledge graphs remains an open question for future investigation.
\section*{Ethics Statement}
The experiments were conducted on publicly available knowledge graphs, eliminating any data privacy concerns. However, it should be noticed that most approaches for generalizable logical entailment are susceptible to adversarial attacks \cite{pmlr-v80-dai18b,ijcai2019p872} and data poisoning \cite{10.1145/3543507.3583203} on knowledge graphs, which may result in unintended outcomes in applications.
\section*{Acknowledgements}
We thank all the anonymous reviewers and meta
reviewers for their valuable comments. The authors of this paper were supported by the ITSP
Platform Research Project (ITS/189/23FP) from
ITC of Hong Kong, SAR, China, and the AoE
(AoE/E-601/24-N), the RIF (R6021-20) and the
GRF (16205322) from RGC of Hong Kong, SAR,
China.
\bibliography{main}
\onecolumn
\newpage
\appendix
\section{Dataset Statistics}
\label{app:benchmark_details}
In this section, we introduce the knowledge graph details and the statistics and query types of our benchmark dataset.

Our benchmark incorporates three widely used knowledge graphs—FB15k, FB15k-237, and NELL995—derived from large-scale KGs to evaluate reasoning capabilities. FB15k and FB15k-237 originate from Freebase, a comprehensive database of structured general knowledge covering domains such as entertainment, sports, and geography, with entities linked by diverse relations (e.g., "born-in," "part-of"). NELL995 is derived from the Never-Ending Language Learning (NELL) system, which extracts structured facts from web text, focusing on real-world entities and relations.

Details of the query statistics for our dataset are presented in Table \ref{tab:num_queries}. For each knowledge graph, we sampled all one-hop projection (\textit{1p}) queries from the training graph and sampled twice as many for the other 22 in-distribution query types. For validation and testing queries, we set the quantities to 8000, 5000, and 4000, respectively, following the convention established in KG literature \cite{ren2020query2box,ren2020beta,DBLP:journals/corr/abs-2302-13114}.

\begin{table}[H]
\small
\centering
\begin{tabular}{ccccc}
\toprule
 \multirow{2}{*}{{Knowledge Graph}}     & \multicolumn{2}{c}{{Training}}    & {Validation} & {Testing} \\\cmidrule{2-5} 
 & 1p & Other Types & All Types          & All Types         \\\midrule
FB15k     & 273,710 & 547,420 & 8,000 & 8,000 \\ \midrule
FB15k-237 & 149,689 & 299,378 & 5,000 & 5,000 \\  \midrule
NELL995  & 107,982 & 215,964 & 4,000 & 4,000 \\ \bottomrule
\end{tabular}
\caption{Number of queries used for each query type in our benchmark.}
\label{tab:num_queries}
\end{table}

Table \ref{tab:id-types} and \ref{tab:ood-types} presented the details of query types in our benchmark. All 32 unseen query types are crafted compositionally based on the 23 seen query types. Table \ref{tab:tree-types} presented the subset of 25 lisp-like compatible query types used in our experiment on query syntax.

\section{Baseline Models}
\label{app:baseline}
In our benchmarking experiment, we compare transformers with five existing methods that are designed particularly for KG query answering:

\begin{itemize}
    \item \textbf{BetaE} \cite{ren2020beta} encodes queries iteratively following the logical operations with Beta distributions. 
    \item \textbf{ConE} \cite{zhang2021cone} represents answer sets as two-dimensional cones to better handle negations. 
    \item \textbf{CQD} \cite{arakelyan2021complex} applies a search-based method that utilizes pre-trained link predictors for inference-time optimization.
    \item \textbf{LMPNN} \cite{wang2023logical} conducts one-hop inferences with pre-trained KGE, and uses message passing to aggregate these local results into global output. 
    \item \textbf{SQE-LSTM} \cite{DBLP:journals/corr/abs-2302-13114} encode the linearized complex query as a sequence with LSTM \cite{hochreiter1997lstm}. 
\end{itemize} 
\newpage
\section{Query Reversion}
The logical formula of five query types before and after reversion is provided in Table \ref{tab:perm-reversion}.
\label{app:reversion}
\begin{table*}[h]

\centering
\small
\begin{tabular}{ccc}
\toprule
Type Name & Before Reversion & After Reversion \\ \midrule
2p &  (r1(s1,e1))\&(r2(e1,f)) &  (r2(e1,f))\&(r1(s1,e1)) \\
3p &  ((r1(s1,e1))\&(r2(e1,e2)))\&(r3(e2,f)) &  ((r3(e2,f))\&(r2(e1,e2)))\&(r1(s1,e1)) \\
ip &  ((r1(s1,e1))\&(r2(s2,e1)))\&(r3(e1,f)) &  ((r3(e1,f))\&(r2(s2,e1)))\&(r1(s1,e1)) \\
pi &  ((r1(s1,e1))\&(r2(e1,f)))\&(r3(s2,f)) &  ((r3(s2,f))\&(r2(e1,f)))\&(r1(s1,e1)) \\
2in &  (r1(s1,f))\&(!(r2(s2,f))) &  (!(r2(s2,f)))\&(r1(s1,f)) \\ \bottomrule
\end{tabular}
\caption{Details of query reversion procedure in the query permutation experiment.}
\label{tab:perm-reversion}
\end{table*}

\setlength{\tabcolsep}{3pt}

\section{Attention Visualization}
\label{app:attn_vis}

\begin{figure}[h]
\begin{center}
\includegraphics[clip,width=370pt]{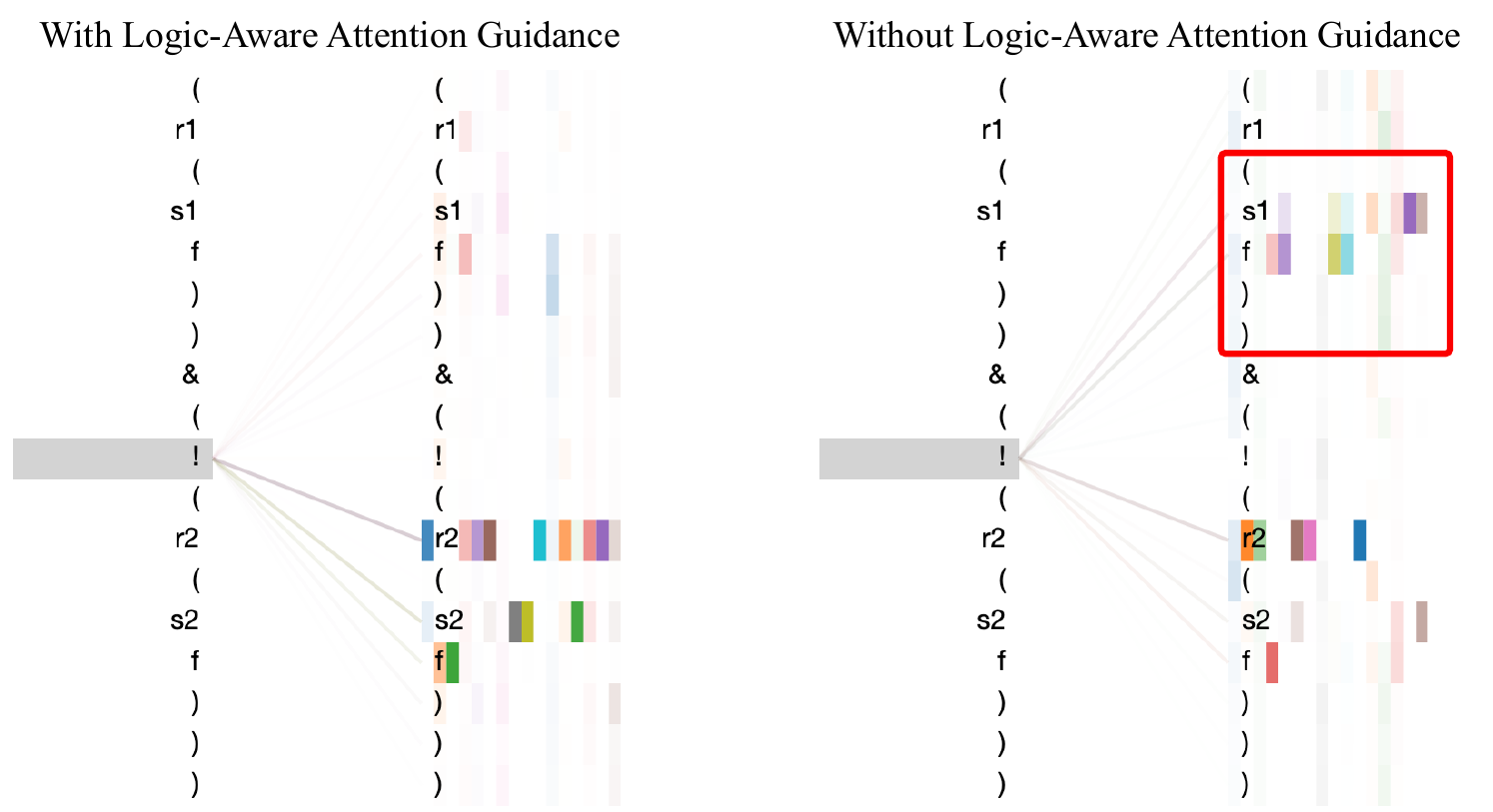}
\end{center}

\caption{An visualization of self-attention when processing \textit{2in} queries. The left and right image is the partial attention weight of TEGA (ours) and Transformer + Relative PE (baseline), respectively.
}
\label{fig:attn_vis}
\end{figure}

Fig.~\ref{fig:attn_vis} provides an example of how attention guidance in TEGA can affect the reasoning process in self-attention. In \textit{2in} queries [\texttt{(r1(s1,f))\&(!(r2(s2,f)))}], the negation operator [\texttt{!}] is applied to calculate the complement of the entity set represented by the second atomic formula [\texttt{(r2(s2,f))}]. Benefited from the attention guidance from inductive biases, TEGA can effectively aggregate the information from the second atomic formula [\texttt{r2/s2/f}] to the negation operator with self-attention. While in baseline (Transformer + Relative PE), the information flow between tokens are scattered and less effective (highlighted in red). As a result, TEGA outperforms Transformer + Relative PE by \textbf{21-26\%} in \textit{2in} queries over three knowledge graphs.

\begin{table*}[]

\centering
\scriptsize
\begin{tabular}{rccccccc}
\toprule
\multicolumn{1}{c}{id} & type name & \multicolumn{1}{c}{mul.} & \multicolumn{1}{c}{cyc.} & \multicolumn{1}{c}{exi.} & \multicolumn{1}{c}{neg.} & depth & logical formula \\ \midrule
0 & 1p &  &  &  &  & 1 & r1(s1,f) \\
1 & 2p &  &  &  &  & 2 & (r1(s1,e1))\&(r2(e1,f)) \\
2 & 3p &  &  &  &  & 3 & (r1(s1,e1))\&(r2(e1,e2))\&(r3(e2,f)) \\
3 & 2i &  &  &  &  & 1 & (r1(s1,f))\&(r2(s2,f)) \\
4 & 3i &  &  &  &  & 1 & (r1(s1,f))\&(r2(s2,f))\&(r3(s3,f)) \\
5 & ip &  &  &  &  & 2 & (r1(s1,e1))\&(r2(s2,e1))\&(r3(e1,f)) \\
6 & pi &  &  &  &  & 2 & (r1(s1,e1))\&(r2(e1,f))\&(r3(s2,f)) \\
7 & 2in &  &  &  & \multicolumn{1}{c}{\checkmark} & 1 & (r1(s1,f))\&(!(r2(s2,f))) \\
8 & 3in &  &  &  & \multicolumn{1}{c}{\checkmark} & 1 & (r1(s1,f))\&(r2(s2,f))\&(!(r3(s3,f))) \\
9 & inp &  &  &  & \multicolumn{1}{c}{\checkmark} & 2 & (r1(s1,e1))\&(!(r2(s2,e1)))\&(r3(e1,f)) \\
10 & pin &  &  &  & \multicolumn{1}{c}{\checkmark} & 2 & (r1(s1,e1))\&(r2(e1,f))\&(!(r3(s2,f))) \\
11 & pni &  &  &  & \multicolumn{1}{c}{\checkmark} & 2 & (r1(s1,e1))\&(!(r2(e1,f)))\&(r3(s2,f)) \\
12 & 2u &  &  &  &  & 1 & (r1(s1,f))|(r2(s2,f)) \\
13 & up &  &  &  &  & 2 & ((r1(s1,e1))|(r2(s2,e1)))\&(r3(e1,f)) \\
14 & 2m & \multicolumn{1}{c}{\checkmark} &  &  &  & 2 & (r1(s1,e1))\&(r2(e1,f))\&(r3(e1,f)) \\
15 & 2nm & \multicolumn{1}{c}{\checkmark} &  &  & \multicolumn{1}{c}{\checkmark} & 2 & (r1(s1,e1))\&(r2(e1,f))\&(!(r3(e1,f))) \\
16 & 3mp & \multicolumn{1}{c}{\checkmark} &  &  &  & 3 & (r1(s1,e1))\&(r2(e1,e2))\&(r3(e2,f))\&(r4(e1,e2)) \\
17 & 3pm & \multicolumn{1}{c}{\checkmark} &  &  &  & 3 & (r1(s1,e1))\&(r2(e1,e2))\&(r3(e2,f))\&(r4(e2,f)) \\
18 & im & \multicolumn{1}{c}{\checkmark} &  &  &  & 2 & (r1(s1,e1))\&(r2(s2,e1))\&(r3(e1,f))\&(r4(e1,f)) \\
19 & 2il &  &  & \multicolumn{1}{c}{\checkmark} &  & 1 & (r1(s1,f))\&(r2(e1,f)) \\
20 & 3il &  &  & \multicolumn{1}{c}{\checkmark} &  & 1 & (r1(s1,f))\&(r2(s2,f))\&(r3(e1,f)) \\
21 & 3c &  & \multicolumn{1}{c}{\checkmark} &  &  & 3 & (r1(s1,e1))\&(r2(e1,f))\&(r3(s2,e2))\&(r4(e2,f))\&(r5(e1,e2)) \\
22 & 3cm & \multicolumn{1}{c}{\checkmark} & \multicolumn{1}{c}{\checkmark} &  &  & 3 & (r1(s1,e1))\&(r2(e1,f))\&(r3(s2,e2))\&(r4(e2,f))\&(r5(e1,e2))\&(r6(e1,f)) \\ \bottomrule
\end{tabular}
\caption{Details of 23 seen query types with their type name, query-level features: \textbf{mul}:multiple relation projection edges, \textbf{cyc}:cycles, \textbf{exi}:existentially quantified variables, \textbf{neg}:negation, \textbf{depth}:longest relation projection chain (reasoning depth), and their logical formula.}
\label{tab:id-types}
\end{table*}

\begin{table*}[]

\centering
\scriptsize
\begin{tabular}{rccccccc}
\toprule
\multicolumn{1}{c}{id} & type name & \multicolumn{1}{c}{mul.} & \multicolumn{1}{c}{cyc.} & \multicolumn{1}{c}{exi.} & neg. & depth & logical formula \\ \midrule
23 & 2pi &  &  &  & \multicolumn{1}{l}{} & 2 & (r1(s1,e1))\&(r2(e1,f))\&(r3(s2,e2))\&(r4(e2,f)) \\
24 & 2pu &  &  &  & \multicolumn{1}{l}{} & 2 & ((r1(s1,e1))\&(r3(e1,f)))|((r2(s2,e2))\&(r4(e2,f))) \\
25 & ui &  &  &  & \multicolumn{1}{l}{} & 2 & ((r1(s1,e1))|(r2(s2,e1)))\&(r3(e1,f))\&(r4(s3,f)) \\
26 & iu &  &  &  & \multicolumn{1}{l}{} & 2 & ((r1(s1,e1))\&(r2(s2,e1))\&(r3(e1,f)))|(r4(s3,f)) \\
27 & upi &  &  &  & \multicolumn{1}{l}{} & 2 & ((r1(s1,e1))|(r2(s2,e1)))\&(r3(e1,f))\&(r4(s3,e2))\&(r5(e2,f)) \\
28 & ipu &  &  &  & \multicolumn{1}{l}{} & 2 & ((r1(s1,e1))\&(r2(s2,e1))\&(r3(e1,f)))|((r4(s3,e2))\&(r5(e2,f))) \\
29 & i2p &  &  &  & \multicolumn{1}{l}{} & 3 & (r1(s1,e1))\&(r2(s2,e1))\&(r3(e1,e2))\&(r4(e2,f)) \\
30 & u2p &  &  &  & \multicolumn{1}{l}{} & 3 & ((r1(s1,e1))|(r2(s2,e1)))\&(r3(e1,e2))\&(r4(e2,f)) \\
31 & 2pin &  &  &  & \checkmark & 2 & (r1(s1,e1))\&(r2(e1,f))\&(r3(s2,e2))\&(r4(e2,f))\&(!(r5(s3,f))) \\
32 & 2pni &  &  &  & \checkmark & 2 & (r1(s1,e1))\&(r2(e1,f))\&(r3(s2,e2))\&(!(r4(e2,f))) \\
33 & pn3i &  &  &  & \checkmark & 2 & (r1(s1,e1))\&(!(r2(e1,f)))\&(r3(s2,f))\&(r4(s3,f)) \\
34 & in2p &  &  &  & \checkmark & 3 & (r1(s1,e1))\&(!(r2(s2,e1)))\&(r3(e1,e2))\&(r4(e2,f)) \\
35 & inu &  &  &  & \checkmark & 2 & ((r1(s1,e1))\&(!(r2(s2,e1)))\&(r3(e1,f)))|(r4(s3,f)) \\
36 & inpu &  &  &  & \checkmark & 2 & ((r1(s1,e1))\&(!(r2(s2,e1)))\&(r3(e1,f)))|((r4(s3,e2))\&(r5(e2,f))) \\
37 & upni &  &  &  & \checkmark & 2 & ((r1(s1,e1))|(r2(s2,e1)))\&(r3(e1,f))\&(r4(s3,e2))\&(!(r5(e2,f))) \\
38 & unpi &  &  &  & \checkmark & 2 & ((r1(s1,e1))|(r2(s2,e1)))\&(!(r3(e1,f)))\&(r4(s3,e2))\&(r5(e2,f)) \\
39 & imp & \multicolumn{1}{c}{\checkmark} &  &  & \multicolumn{1}{l}{} & 3 & (r1(s1,e1))\&(r2(s2,e1))\&(r3(e1,e2))\&(r4(e2,f))\&(r5(e1,e2)) \\
40 & ipm & \multicolumn{1}{c}{\checkmark} &  &  & \multicolumn{1}{l}{} & 3 & (r1(s1,e1))\&(r2(s2,e1))\&(r3(e1,e2))\&(r4(e2,f))\&(r5(e2,f)) \\
41 & 3im & \multicolumn{1}{c}{\checkmark} &  &  & \multicolumn{1}{l}{} & 2 & (r1(s1,e1))\&(r2(s2,e1))\&(r3(s3,e1))\&(r4(e1,f))\&(r5(e1,f)) \\
42 & pil &  &  & \multicolumn{1}{c}{\checkmark} & \multicolumn{1}{l}{} & 2 & (r1(s1,e1))\&(r2(e1,f))\&(r3(e2,f)) \\
43 & ilp &  &  & \multicolumn{1}{c}{\checkmark} & \multicolumn{1}{l}{} & 2 & (r1(s1,e1))\&(r2(e2,e1))\&(r3(e1,f)) \\
44 & p3il &  &  & \multicolumn{1}{c}{\checkmark} & \multicolumn{1}{l}{} & 2 & (r1(s1,e1))\&(r2(e1,f))\&(r3(s2,e2))\&(r4(e2,f))\&(r5(e3,f)) \\
45 & i3c &  & \multicolumn{1}{c}{\checkmark} &  & \multicolumn{1}{l}{} & 3 & (r1(s1,e1))\&(r2(s2,e1))\&(r3(e1,f))\&(r4(s3,e2))\&(r5(e2,f))\&(r6(e1,e2)) \\
46 & i3cm & \multicolumn{1}{c}{\checkmark} & \multicolumn{1}{c}{\checkmark} &  & \multicolumn{1}{l}{} & 3 & (r1(s1,e1))\&(r2(s2,e1))\&(r3(e1,f))\&(r4(s3,e2))\&(r5(e2,f))\&(r6(e1,e2))\&(r7(e1,f)) \\
47 & 3inl &  &  & \multicolumn{1}{c}{\checkmark} & \checkmark & 1 & (r1(s1,f))\&(!(r2(s2,f)))\&(r3(e1,f)) \\
48 & pinl &  &  & \multicolumn{1}{c}{\checkmark} & \checkmark & 2 & (r1(s1,e1))\&(r2(e1,f))\&(!(r3(s2,f)))\&(r4(e2,f)) \\
49 & inm & \multicolumn{1}{c}{\checkmark} &  &  & \checkmark & 2 & (r1(s1,e1))\&(!(r2(s2,e1)))\&(r3(e1,f))\&(r4(e1,f)) \\
50 & inmp & \multicolumn{1}{c}{\checkmark} &  &  & \checkmark & 3 & (r1(s1,e1))\&(!(r2(s2,e1)))\&(r3(e1,e2))\&(r4(e2,f))\&(r5(e1,e2)) \\
51 & inpm & \multicolumn{1}{c}{\checkmark} &  &  & \checkmark & 3 & (r1(s1,e1))\&(!(r2(s2,e1)))\&(r3(e1,f))\&(r4(e2,f))\&(r5(e2,f)) \\
52 & 3nmp & \multicolumn{1}{c}{\checkmark} &  &  & \checkmark & 3 & (r1(s1,e1))\&(r2(e1,e2))\&(r3(e2,f))\&(!(r4(e1,e2))) \\
53 & 3cn &  & \multicolumn{1}{c}{\checkmark} &  & \checkmark & 3 & (r1(s1,e1))\&(r2(e1,f))\&(r3(s2,e2))\&(!(r4(e2,f)))\&(r5(e1,e2)) \\
54 & 3cnm & \multicolumn{1}{c}{\checkmark} & \multicolumn{1}{c}{\checkmark} &  & \checkmark & 3 & (r1(s1,e1))\&(r2(e1,f))\&(r3(s2,e2))\&(r4(e2,f))\&(r5(e1,e2))\&(!(r6(e1,f))) \\ \bottomrule
\end{tabular}
\caption{Details of 32 unseen query types with their type name, query-level features: \textbf{mul}:multiple relation projection edges, \textbf{cyc}:cycles, \textbf{exi}:existentially quantified variables, \textbf{neg}:negation, \textbf{depth}:longest relation projection chain (reasoning depth), and their logical formula.}
\label{tab:ood-types}
\end{table*}

\begin{table*}[]

\scriptsize
\centering
\begin{tabular}{ccc}
\toprule
\multicolumn{3}{c}{Seen Query Types (13 types)} \\ \midrule
Type Name & EFO Syntax & Lisp-Like Syntax \\ \midrule
1p & r1(s1,f) & (p,(r1),(s1)) \\
2p & (r1(s1,e1))\&(r2(e1,f)) & (p,(r2),(p,(r1),(s1))) \\
3p & (r1(s1,e1))\&(r2(e1,e2))\&(r3(e2,f)) & (p,(r3),(p,(r2),(p,(r1),(s1)))) \\
2i & (r1(s1,f))\&(r2(s2,f)) & (i,(p,(r1),(s1)),(p,(r2),(s2))) \\
3i & (r1(s1,f))\&(r2(s2,f))\&(r3(s3,f)) & (i,(p,(r1),(s1)),(p,(r2),(s2)),(p,(r3),(s3))) \\
ip & (r1(s1,e1))\&(r2(s2,e1))\&(r3(e1,f)) & (p,(r3),(i,(p,(r1),(s1)),(p,(r2),(s2)))) \\
pi & (r1(s1,e1))\&(r2(e1,f))\&(r3(s2,f)) & (i,(p,(r2),(p,(r1),(s1))),(p,(r3),(s2))) \\
2in & (r1(s1,f))\&(!(r2(s2,f))) & (i,(p,(r1),(s1)),(n,(p,(r2),(s2)))) \\
3in & (r1(s1,f))\&(r2(s2,f))\&(!(r3(s3,f))) & (i,(p,(r1),(s1)),(p,(r2),(s2)),(n,(p,(r3),(s3)))) \\
inp & (r1(s1,e1))\&(!(r2(s2,e1)))\&(r3(e1,f)) & (p,(r3),(i,(p,(r1),(s1)),(n,(p,(r2),(s2))))) \\
pin & (r1(s1,e1))\&(r2(e1,f))\&(!(r3(s2,f))) & (i,(p,(r2),(p,(r1),(s1))),(n,(p,(r3),(s2)))) \\
2u & (r1(s1,f))|(r2(s2,f)) & (u,(p,(r1),(s1)),(p,(r2),(s2))) \\
up & ((r1(s1,e1))|(r2(s2,e1)))\&(r3(e1,f)) & (p,(r3),(u,(p,(r1),(s1)),(p,(r2),(s2)))) \\ \midrule
\multicolumn{3}{c}{Unseen Query Types (12 types)} \\ \midrule
Type Name & EFO Syntax & Lisp-Like Syntax \\ \midrule
2pi & (r1(s1,e1))\&(r2(e1,f))\&(r3(s2,e2))\&(r4(e2,f)) & (i,(p,(r2),(p,(r1),(s1))),(p,(r4),(p,(r3),(s2)))) \\
2pu & ((r1(s1,e1))\&(r3(e1,f)))|((r2(s2,e2))\&(r4(e2,f))) & (u,(p,(r3),(p,(r1),(s1))),(p,(r4),(p,(r2),(s2)))) \\
ui & ((r1(s1,e1))|(r2(s2,e1)))\&(r3(e1,f))\&(r4(s3,f)) & (i,(p,(r4),(s3)),(p,(r3),(u,(p,(r1),(s1)),(p,(r2),(s2))))) \\
iu & ((r1(s1,e1))\&(r2(s2,e1))\&(r3(e1,f)))|(r4(s3,f)) & (u,(p,(r4),(s3)),(p,(r3),(i,(p,(r1),(s1)),(p,(r2),(s2))))) \\
upi & ((r1(s1,e1))|(r2(s2,e1)))\&(r3(e1,f))\&(r4(s3,e2))\&(r5(e2,f)) & (i,(p,(r5),(p,(r4),(s3))),(p,(r3),(u,(p,(r1),(s1)),(p,(r2),(s2))))) \\
ipu & ((r1(s1,e1))\&(r2(s2,e1))\&(r3(e1,f)))|((r4(s3,e2))\&(r5(e2,f))) & (u,(p,(r5),(p,(r4),(s3))),(p,(r3),(i,(p,(r1),(s1)),(p,(r2),(s2))))) \\
i2p & (r1(s1,e1))\&(r2(s2,e1))\&(r3(e1,e2))\&(r4(e2,f)) & (p,(r4),(p,(r3),(i,(p,(r1),(s1)),(p,(r2),(s2))))) \\
u2p & ((r1(s1,e1))|(r2(s2,e1)))\&(r3(e1,e2))\&(r4(e2,f)) & (p,(r4),(p,(r3),(u,(p,(r1),(s1)),(p,(r2),(s2))))) \\
2pin & (r1(s1,e1))\&(r2(e1,f))\&(r3(s2,e2))\&(r4(e2,f))\&(!(r5(s3,f))) & (i,(i,(p,(r4),(p,(r3),(s2))),(p,(r2),(p,(r1),(s1)))),(n,(p,(r5),(s3)))) \\
in2p & (r1(s1,e1))\&(!(r2(s2,e1)))\&(r3(e1,e2))\&(r4(e2,f)) & (p,(r4),(p,(r3),(i,(p,(r1),(s1)),(n,(p,(r2),(s2)))))) \\
inu & ((r1(s1,e1))\&(!(r2(s2,e1)))\&(r3(e1,f)))|(r4(s3,f)) & (u,(p,(r4),(s3)),(p,(r3),(i,(p,(r1),(s1)),(n,(p,(r2),(s2)))))) \\
inpu & ((r1(s1,e1))\&(!(r2(s2,e1)))\&(r3(e1,f)))|((r4(s3,e2))\&(r5(e2,f))) & (u,(p,(r5),(p,(r4),(s3))),(p,(r3),(i,(p,(r1),(s1)),(n,(p,(r2),(s2)))))) \\ \bottomrule
\end{tabular}
\caption{Details of 25 query types (compatible in both Lisp-like and EFO syntax) that we used as a subset for exploration on query syntax.}
\label{tab:tree-types}
\end{table*}
\newpage
\section{Knowledge Graph Statistics}
Detailed statistics of the knowledge graphs selected are presented in Table \ref{tab:KG_details}.
\begin{table}[h]

\begin{center}
\small

\begin{tabular}{lcccccc} 
\toprule
Dataset & Relations & Entities & Training & Validation & Testing & All Edges\\
\midrule
FB15k & 1,345 & 14,951 & 483,142 & 50,000 & 59,071 & 592,213\\
FB15k-237 & 237  & 14,505& 272,115 & 17,526 & 20,438 & 310,079\\
NELL995 & 200  & 63,361& 114,213 & 14,324 & 14,267 & 142,804\\
\bottomrule
\end{tabular}
\caption{Details of three knowledge graphs used for the experiments, and their separation standard for training, validation, and testing edges according to \cite{ren2020beta}.}
\label{tab:KG_details}

\end{center}
\end{table}

\setlength{\tabcolsep}{6pt}
\section{Query Graph Definition}
\label{app:query_graph}
We provide query graph for each query types in our dataset. Query graphs can be utilized for graph-augmented methods, and have been applied in the two inductive biases discussed in Sec. \ref{sec:transformer_architecture}. Here we provide the definition:

For each atomic formula or its negation $\alpha = r(h, t)$ or $\neg r(h, t)$ in a conjunctive formula $c$, we have $\{(h,r),(r,t)\}\in \mathcal{G}_c$ or $\{(h,r),(r,n), (n,t)\}\in \mathcal{G}_c$, where $n$ denotes the negation node in the conjunctive query graph $\mathcal{G}_c$. By our definition of query, all conjunctive query graphs have exactly one node as a free variable. For queries that represent the disjunction of multiple conjunctive formulas, we replace all free variable nodes in every conjunctive query graph with a single union node $u$, and connect it to a final free variable node $f$, with $\{(u,f)\}\in \mathcal{G}_d$, where $\mathcal{G}_d$ is the disjunctive query graph.
\newpage
\section{Evaluation Details}
\label{app:evaluation_details}
For each knowledge graphs, we separate their edges into training, validation, and testing edges with a ratio of approximately 8:1:1, as shown in Table \ref{tab:KG_details}. We construct three graphs, training graph $\mathcal{G}_{train}$, validation graph $\mathcal{G}_{valid}$, and testing graph $\mathcal{G}_{test}$ with training edges, training+validation edges, and training+validation+testing edges, respectively. 
 
We adopt the mean reciprocal rank (MRR) as our evaluation metric, following the calculation presented in Equation \ref{equa:mrri} and \ref{equa:mrro}. It is important to note that for all testing queries, the answers in $A(\mathcal{G}_{valid},q) - A(\mathcal{G}_{train},q)$ are excluded from $\mathcal{A}_{ood}$ to ensure fairness. Consequently, we define $\mathcal{A}_{id} = A(\mathcal{G}_{train},q)$ and $\mathcal{A}_{ood} = A(\mathcal{G}_{test},q) - A(\mathcal{G}_{valid},q)$ for our evaluation of ID (K) and OOD (K), respectively. In each table regarding experimental results, the scores below ID (Q) (i.e., either MRR of ID (K) or OOD (K)) represent the average scores among all seen query types. While the scores below OOD (Q) represent the average scores among all unseen query types. To connect with literature in KG query answering \cite{DBLP:journals/corr/abs-2302-13114,sun2021faithful}, the performances in ID (K) and OOD (K) are also referred to as \textit{faithfulness} and \textit{knowledge inference capability}, respectively.

\begin{align}
    \label{equa:mrri}
    MRR_{id(k)} &= \frac{\sum_{v \in \mathcal{A}_{id}}}{|\mathcal{A}_{id}|\texttt{rank}(v)} \\
    MRR_{ood(k)} &= \frac{\sum_{v \in \mathcal{A}_{ood}}}{|\mathcal{A}_{ood}|\texttt{rank}(v)}
    \label{equa:mrro}
\end{align}

\section{Transformer Architecture Comparison}
\label{app:transformer_arch}

We investigate the impact of different transformer architectures on complex query answering performance. Table \ref{tab:transformer_arch} compares three architectural variants: encoder-only (our main approach), decoder-only, and encoder-decoder configurations. The results demonstrate that the encoder-only architecture with relative positional encoding achieves the best overall performance across datasets. 
\begin{table*}[h]
\centering
\small
\begin{tabular}{clccccccc}
\toprule
\multirow{2}{*}{Dataset} & \multicolumn{1}{c}{\multirow{2}{*}{Architecture}} & \multirow{2}{*}{PE Type} & \multicolumn{2}{c}{ID (Q)} & \multicolumn{2}{c}{OOD (Q)} & \multicolumn{2}{c}{All Queries} \\\cmidrule{4-9}
 & \multicolumn{1}{c}{} &  & ID (K) & OOD (K) & ID (K) & OOD (K) & ID (K) & OOD (K) \\ \midrule
\multirow{4}{*}{FB15k} 
 & Encoder-only & Absolute PE & 46.9 & {\ul 31.9} & 21.8 & 13.2 & 32.3 & 21.0 \\
 & Encoder-only & Relative PE & \textbf{48.1} & \textbf{32.3} & \textbf{35.4} & \textbf{21.5} & \textbf{40.7} & \textbf{26.0} \\
 & Decoder-only & Absolute PE & {\ul 47.9} & {\ul 31.9} & 24.3 & 14.7 & 34.2 & 21.9 \\
 & Encoder-Decoder & Relative PE & 40.1 & 28.5 & {\ul 32.0} & {\ul 19.8} & {\ul 35.4} & {\ul 23.4} \\ \midrule
\multirow{4}{*}{FB15k-237} 
 & Encoder-only & Absolute PE & 54.4 & \textbf{20.0} & 20.8 & 9.0 & 34.9 & 13.6 \\
 & Encoder-only & Relative PE & {\ul 54.8} & \textbf{20.0} & \textbf{37.7} & \textbf{15.8} & \textbf{44.8} & \textbf{17.6} \\
 & Decoder-only & Absolute PE & \textbf{55.2} & 19.9 & 21.9 & 9.8 &  35.8 & 14.0 \\
 & Encoder-Decoder & Relative PE & 44.5 & 19.9 & {\ul 33.5} & \textbf{15.8} & {\ul 38.1} & {\ul 17.5} \\ \midrule
\multirow{4}{*}{NELL995} 
 & Encoder-only & Absolute PE & \textbf{68.8} & {\ul 18.9} & 25.8 & 8.3 & 43.8 & 12.7 \\
 & Encoder-only & Relative PE & {\ul 68.0} & 18.7 & \textbf{49.8} & {\ul 13.4} & \textbf{57.4} & {\ul 15.6} \\
 & Decoder-only & Absolute PE & 65.6 & 18.6 & 28.0 & 8.9 & 43.7 & 13.0 \\
 & Encoder-Decoder & Relative PE & 65.5 & \textbf{19.7} & {\ul 48.7} & \textbf{13.7} & {\ul 55.7} & \textbf{16.2} \\ \bottomrule
\end{tabular}
\caption{Comparison of different transformer architectures (in MRR\%).}
\label{tab:transformer_arch}
\end{table*}

\newpage
\section{Technical Details}
\label{app:technical_details}
In this section, we describe our experiment setting in more details.

\textbf{Computational Resource}: All transformers are trained on four NVIDIA A100 GPUs for two days, with a batch size of 1024. As shown in Table \ref{tab:tega_effi}, TEGA achieves computational efficiency comparable to that of the baseline models.

\textbf{Model Configuration}: Table \ref{tab:scale} illustrated the impact of transformer size to query answering performances. We selected a three-layer configuration for all Transformer models, for its effectiveness and comparable parameter size to other baseline methods (as shown in Table \ref{tab:num-params}). To train the Transformer models, we employed the label smoothing loss \cite{szegedy2015rethinking} with a smoothing factor of $\epsilon = 0.1$, which helps to regularize the model and improve generalization. The learning rate was set to 0.0001, and a warm-up schedule of 1000 steps was applied to gradually increase the learning rate during the initial phase of training, allowing the model to adapt to the task and stabilize the optimization process. All Transformer models and baselines were configured with an embedding size of 400 dimensions. Regarding transformer encoder with absolute and relative PE, we employ the \texttt{BertModel} from the \texttt{transformers} library. While for disentangled PE, we use \texttt{DebertaModel} with the same size and configurations. Note that apart from disentangled PE, Deberta also proposed a mask decoder during the pre-training stage. However, as we train all models solely on logical queries without utilizing any pre-trained model weights, the impact of the mask decoder is negligible.

\textbf{Pooling Strategy}: For free-variable pooling, we experimented with two pooling settings: sum pooling and mean pooling, with results presented in Table \ref{tab:pooling}. The results show that sum pooling over free variables is the better setting, which is also consistent with findings in GNN literature \cite{xu2019powerful}.
\newpage 
\begin{table*}[t]
\small
\centering
\begin{tabular}{lc}
\toprule
\multicolumn{1}{c}{Models} & Inference Time (ms) \\ \midrule
Transformer (Absolute PE) & 43.2 \\
Transformer (Disentangled PE) & 47.3 \\
Transformer (Relative PE) & 44.7 \\
+Adjacency Matrices Masking & 46.8 \\
+Direct Distance Encoding & 44.9 \\
TEGA (ours) & 52.1 \\ \bottomrule
\end{tabular}
\caption{Computational efficiency of TEGA compared with baselines.}
\label{tab:tega_effi}
\end{table*}
\begin{table*}[]
\centering
\small
\begin{tabular}{@{}cc@{}}
\toprule
Model & Number of Trainable Parameters (in Millions) \\ \midrule
BetaE & 13.7 \\
ConE & 18.9 \\
LMPNN & 11.5 \\
SQE-LSTM & 8.8 \\
Trans.+Relative PE & 14.7 \\
\textbf{TEGA (ours)} & \textbf{15.7} \\ \bottomrule
\end{tabular}
\caption{Parameter size comparison between different query answering approaches, with three layers of self-attention in transformer models.}
\label{tab:num-params}
\end{table*}

\begin{table*}[]
\centering
\small
\begin{tabular}{@{}cccccccc@{}}
\toprule
\multirow{2}{*}{Model} & \multirow{2}{*}{\# Attn. layers} & \multicolumn{2}{c}{ID(Q)} & \multicolumn{2}{c}{OOD(Q)} & \multicolumn{2}{c}{All Queries} \\ \cmidrule(l){3-8} 
 &  & ID(K) & OOD(K) & ID(K) & OOD(K) & ID(K) & OOD(K) \\ \midrule
\multirow{3}{*}{Trans.+Absolute PE} & 3 & 54.4 & {\ul 20.0} & 20.8 & 9.0 & 34.9 & 13.6 \\
 & 6 & \textbf{57.2} & \textbf{20.3} & 23.0 & 9.6 & 37.3 & 14.1 \\
 & 12 & 48.9 & 19.8 & 24.4 & 11.4 & 34.6 & 14.9 \\ \midrule
\multirow{3}{*}{Trans.+Relative PE} & 3 & 54.8 & {\ul 20.0} & {\ul 37.7} & {\ul 15.8} & {\ul 44.8} & {\ul 17.6} \\
 & 6 & {\ul 56.7} & {\ul 20.0} & \textbf{38.0} & \textbf{16.0} & \textbf{45.8} & \textbf{17.7} \\
 & 12 & 48.6 & 19.9 & 35.0 & 15.7 & 40.7 & 17.4 \\ \bottomrule
\end{tabular}
\caption{Experimental results on transformers with different numbers of attention layers in the FB15K-237 dataset.}
\label{tab:scale}
\end{table*}

\begin{table*}[t]

\centering
\small
\begin{tabular}{cccccccc}
\toprule
\multirow{2}{*}{Dataset} & \multirow{2}{*}{Pooling} & \multicolumn{2}{c}{ID (Q)} & \multicolumn{2}{c}{OOD (Q)} & \multicolumn{2}{c}{All Queries} \\\cmidrule{3-8}
 &  & ID (K) & OOD (K) & ID (K) & OOD (K) & ID (K) & OOD (K) \\ \midrule
\multirow{2}{*}{FB15k} & Mean & 52.9 & 33.5 & 37.2 & 21.6 & 43.7 & 26.6 \\
 & Sum & \textbf{55.0} & \textbf{34.1} & \textbf{38.8} & \textbf{22.4} & \textbf{45.6} & \textbf{27.3} \\ \midrule
\multirow{2}{*}{FB15k-237} & Mean & 62.9 & 20.0 & 41.6 & 15.6 & 50.5 & 17.4 \\
 & Sum & \textbf{64.3} & \textbf{20.1} & \textbf{42.6} & \textbf{16.0} & \textbf{51.7} & \textbf{17.7} \\ \bottomrule
\end{tabular}
\caption{Performance comparison between sum pooling and mean pooling strategies on TEGA.}
\label{tab:pooling}
\end{table*}

\end{document}